\documentclass[11pt,a4paper]{article}
\usepackage[margin=25mm]{geometry}
\usepackage[T1]{fontenc}
\usepackage{lmodern,amsmath,amssymb,booktabs,tabularx,longtable,microtype}
\usepackage{xcolor,tikz,pgfplots,caption}
\usetikzlibrary{arrows.meta,positioning,calc}
\pgfplotsset{compat=1.18}
\definecolor{StudyBlue}{HTML}{286AA6}
\definecolor{StudyTeal}{HTML}{218F86}
\definecolor{StudyOrange}{HTML}{C97632}
\definecolor{StudyGray}{HTML}{7B8290}
\usepackage[colorlinks=true,linkcolor=StudyBlue,urlcolor=StudyBlue,citecolor=StudyBlue]{hyperref}
\hypersetup{pdftitle={What Does History Add? Calendar Controls and Information-Timing Audits for University Trajectory Forecasting}}
\title{What Does History Add?\\Calendar Controls and Information-Timing Audits for University Trajectory Forecasting}
\author{Zepei Cao\\East China Normal University}
\date{15 October 2026}
\begin{document}
\maketitle

\begin{abstract}
University planning systems may benefit from prior observations, but a comparison with a current-state predictor can conflate historical information with additional calendar features. We examine this distinction in a reproducible, rule-based university trajectory simulator with 11 state coordinates, planned activities, resource constraints, and stochastic execution. The study reconstructs 16,750 routes from 3,350 simulated profiles, comprising 804,000 monthly records across training, validation, and two family-level evaluation sets. Seventeen regularized linear specifications and a persistence baseline separate current plans, past-state summaries, past-plan summaries, and history-length indicators. On the original 40-family evaluation, six-month history reduces six-step normalized RMSE from 0.159234 to 0.151522. However, its advantage over a history-length-only control is only $-0.000895$, with a paired family-bootstrap 95\% interval of $[-0.001957,0.000248]$. Frozen weights evaluated on 30 previously unused families show the same pattern: 0.156480 versus 0.150074 against the current-state baseline, but a length-controlled difference of $-0.000558$ with 95\% interval $[-0.001783,0.000728]$. An exploratory loss decomposition identifies improvements in preparation indices and funds, while fatigue accounts for 97.80\% of the history model's original six-step normalized squared error and has an uncertain length-controlled benefit. Moreover, 28.20\% of its six-step fatigue predictions violate the simulator's actual $[5,90]$ support. The results support an audited distinction between history summaries, clock information, and state-specific predictive value. They concern conditional simulator prediction given recorded future plans; they do not establish real-student forecasting, causal action effects, autonomous planning, or language-model memory performance.
\end{abstract}

\section{Introduction}
University planning spans several interacting processes: academic preparation, projects, career exploration, limited discretionary time, and available funds. A system that represents only the latest state may discard useful information about recent progress or repeated activity. Conversely, a system with extensive history can appear stronger because it receives additional indicators of elapsed time, because evaluation supplies later plans unavailable at the forecast origin, or because an aggregate score conceals differences among outcomes. These possibilities must be distinguished before attributing an improvement to memory.

We study a deliberately limited question: what additional predictive value do explicitly defined summaries of earlier states and plans provide, beyond the current state, nominal context, and current recorded plan? A rule-based simulator makes the information order and target-generation mechanism inspectable. It also limits the substantive interpretation: its internal preparation and fatigue indices are designed quantities, not validated measurements of actual students. This setting is useful for examining evaluation logic and conditional prediction, while offering no substitute for empirical educational validation.

The key control is inexpensive. A windowed-history model receives an available-history count and an indicator that history is nonempty. Both are deterministic functions of the calendar. Adding them to a linear model changes its calendar basis even when no earlier observation is supplied. We therefore compare each relevant history model with a length-only counterpart, alongside the conventional current-state baseline. The resulting distinction matters: a favorable comparison against the latter need not imply a favorable comparison against the former.

Our contribution is a task and evaluation audit rather than a new recurrent architecture. We formalize the simulator's prediction-time contract, compare 17 matched Ridge specifications, preserve scenario-family dependence in evaluation, and inspect target-specific error and output support. An exposure audit found that 22 of the original 40 evaluation families had appeared in earlier local tasks. We retain those preselected results transparently and evaluate locked weights on a separately selected 30-family set that excludes all identified previous uses. The substantive finding is conditional: historical summaries improve several preparation and funds predictions, but the prespecified six-month summary's six-step aggregate advantage over its clock-aware control remains uncertain in both evaluation sets. All-history summaries retain a clearer advantage over their own count-all control. This distinction is more informative than a general claim that longer memory improves university planning.

\section{Related Work}
Recurrent educational state estimation is established. Deep Knowledge Tracing models student coursework interactions to predict subsequent performance~\cite{piech2015dkt}. CSEAL combines LSTM knowledge tracing, prerequisite navigation, and actor--critic learning-path recommendation~\cite{liu2019cseal}. Its main learning-gain experiments use both a rule-based simulator and a DKT-based simulator, with explicit discussion of their limitations. Our domain includes several preparation dimensions, fatigue, and funds, but the combination of dynamic state, history, constraints, and educational recommendation is prior work. We do not claim a first education world model or a new theory of student development.

Learned dynamics can support action selection when evaluated in the environment in which actions are executed. PlaNet connects latent dynamics and multi-step prediction with online planning and control~\cite{hafner2019planet}. Sequential predictions and policy choices can also alter the distribution of subsequent observations, as analyzed in DAgger~\cite{ross2011dagger}; MOPO addresses distribution shift in model-based offline policy optimization~\cite{yu2020mopo}. These works motivate separating conditional prediction from policy evaluation. Our supplied-plan experiment trains and evaluates state predictors, without learning an autonomous policy or optimizing an offline return.

Architecture terminology also requires care. GRU's reset and update gates originate in established recurrent-network work~\cite{cho2014gru}; neither a recurrent hidden vector nor a history summary establishes human-like long-term memory. This study instead uses transparent linear predictors to isolate declared information channels. Its conclusions apply to these representations and this data-generating mechanism, rather than to all possible memory architectures.

Finally, feature legitimacy, independence assumptions, and the scientific target of prediction are central to reproducibility~\cite{kapoor2023leakage}. A task can be valid conditional on supplied information while being unsuitable for a different deployment-time claim. Real educational outcomes require additional evidence. For example, a real-world educational reinforcement-learning study found no significant reading-competence difference from its expert baseline despite improvements within both groups~\cite{riedmann2025realworld}. Simulator accuracy and educational effectiveness should therefore remain separate conclusions.

\section{Data Provenance and Information Contract}
\subsection{Inspectable simulator records}
The source archive contains 30,000 simulated starting profiles, five policy-generated routes per profile, and 48 months per route. Its 150,000 routes are not 150,000 independently observed students. The generator identifies profiles and monthly histories as synthetic, marks numerical coefficients as uncalibrated design assumptions, and provides deterministic reconstruction under its frozen version and seed. Public factual references inform scenario vocabulary and observable endpoint fragments; they do not supply complete longitudinal observations for these generated profiles.

The predictor targets nine internal preparation indices, an internal fatigue index, and a discretionary-funds balance in scenario yuan. These quantities follow explicit transition equations and stochastic execution. They are not calibrated probabilities of admission or employment, psychometric scores, or verified personal financial histories. Source family groups contain 25 profile configurations. Each profile's five routes share its starting conditions, and related configurations within a family must remain together when splitting and resampling.

We reconstruct 104 families for the expanded experiment: 48 training, 16 validation, and 40 evaluation families. This yields 1,200 training profiles, 400 validation profiles, and 1,000 original evaluation profiles, or 6,000, 2,000, and 5,000 routes, respectively. The separate 30-family evaluation adds 750 profiles and 3,750 routes. Thus the study processes 134 distinct families, 3,350 profiles, 16,750 routes, and 804,000 monthly records in total. Only the two evaluation sets are used to report held-out prediction performance; the total count includes training and validation.

\subsection{State, event, plan, and execution order}
\begin{figure}[htbp]
\centering
\begin{tikzpicture}[every node/.style={font=\scriptsize}, box/.style={draw,rounded corners,text width=24mm,align=center,minimum height=13mm}, >={Stealth}]
\node[box,draw=StudyBlue] (s) at (0,0) {Observed prefix\\and pre-event state};
\node[box,draw=StudyOrange] (e) at (2.7,0) {Same-month\\events};
\node[box,draw=StudyBlue] (a) at (5.4,0) {Saved plan\\formed after events};
\node[box,draw=StudyGray] (z) at (8.1,0) {Execution and\\hidden progress};
\node[box,draw=StudyTeal] (y) at (10.8,0) {Next recorded\\state};
\draw[->] (s)--(e);\draw[->] (e)--(a);\draw[->] (a)--(z);\draw[->] (z)--(y);
\draw[->,StudyOrange] (e.north) to[bend left=20] (y.north);
\node[text width=11cm,align=center,above=13mm of a] {Events can influence both the saved plan and the resulting state.};
\end{tikzpicture}
\caption{Information timing in the source simulator. One-step prediction is conditional on an already formed saved plan. Recursive evaluation additionally supplies future recorded plans, but not future observed state vectors.}
\label{fig:clock}
\end{figure}
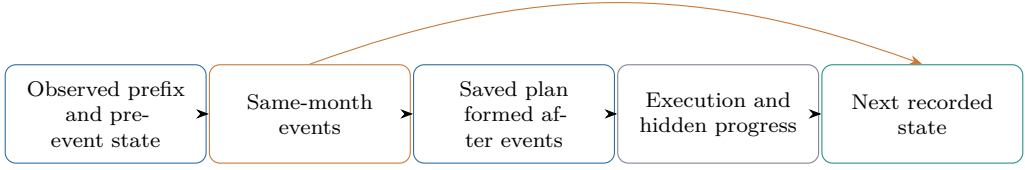

Figure~\ref{fig:clock} depicts the source ordering. For monthly record $t$, the simulator first stores a pre-event state $x_t$. It then samples events and updates resources, forms the saved plan $a_t$, executes activities stochastically, and records the resulting state $y_t$. Consecutive records satisfy $y_t=x_{t+1}$ for the selected core coordinates. Nominal context $c_t$ contains fixed profile resources, scenario descriptors, and deterministic calendar features.

Our conditional input consists of $x_t$, $c_t$, the saved plan when enabled, and permitted earlier observations. Actual hours, realized spending, event fields, execution status, narrative rationale, identifiers, and target states are excluded. Nevertheless, $a_t$ is formed after the current event and resource update. It can consequently carry predictive information about those changes. A comparison with and without $a_t$ estimates the value of this recorded information under the declared task; it does not identify the effect of forcing an action.

Multi-step evaluation supplies recorded future plans. Those plans were originally formed using later realized states and events. Consequently, the rollout is retrospective conditional state prediction, not a forecast available from all information at the initial month. Excluding future true state vectors from recursion does not make these plans available prospectively. We use the word ``forecasting'' only with this explicit conditioning qualification and make no claim of autonomous replanning.

\section{Prediction Models and Temporal Summaries}
\subsection{Strictly past summary features}
Let $t\in\{0,\ldots,47\}$ and let $k_t(w)=\min(t,w)$ for a finite window $w$, or $k_t(\infty)=t$ for all available history. For $k_t>0$, we define
\begin{align}
 m_t^{x}(w)&=\frac{1}{k_t}\sum_{j=t-k_t}^{t-1}x_j,\\
 m_t^{\Delta}(w)&=\frac{x_t-x_{t-k_t}}{k_t},\\
 m_t^{a}(w)&=\frac{1}{k_t}\sum_{j=t-k_t}^{t-1}a_j.
\end{align}
The past-state mean excludes $x_t$, and the past-plan mean excludes $a_t$. The change summary is the mean of completed state differences ending at the current observed state. Its telescoping form shows that it is an endpoint slope, not a representation of every transition or its order. At $t=0$, all summaries are zero. Summary models additionally receive
\begin{equation}
 \ell_t(w)=\left(k_t(w)/48,\;\mathbf{1}\{t>0\}\right).
\end{equation}

The baseline feature vector is $[x_t,c_t,a_t]$. The full windowed vector appends all three summaries and $\ell_t(w)$. We evaluate windows of 1, 3, 6, and 12 months and all available history; component-only and component-removal specifications; current-plan removal; and two length-only controls. A length-only model appends $\ell_t(w)$ but no historical states or plans. Although the baseline already contains a month fraction and seasonal indicators, finite-window $\ell_t(w)$ introduces a clipped linear function and an initial-record indicator. It therefore supplies a different calendar basis to a linear model.

The baseline has 94 input coordinates, the full history model 164, and the length-only control 96. For a finite window, the count supplies a saturating calendar function. For all history, $k_t/48=t/48$ is affine-equivalent to the baseline month fraction $(t+1)/48$, while the nonempty indicator still adds an initial-record distinction. Duplicating an affine-equivalent column can also change the Ridge penalty geometry. Thus these controls expose specific clock and representation effects rather than perfectly removing every calendar or capacity difference. Comparisons concern additional summary information within the specified regularized linear model family. They do not isolate an architecture-independent value of memory. Means discard temporal order, and the all-history slope can be reconstructed from the current and initial states with elapsed time.

\subsection{Shared regularized regression}
All 17 learned specifications use the same Ridge fitting procedure. Features and target changes $y_t-x_t$ are standardized using training records only. An unpenalized intercept accompanies a squared-error objective with an $L_2$ penalty on coefficients. Each specification independently selects its penalty from $\{0.01,0.1,1,10,100,1000\}$ by the validation one-step score; test records never select a penalty or window. Predictions are obtained by adding the predicted change to the current state. We also report persistence, which retains the initial state throughout a rollout. Thus the original evaluation contains 17 learned Ridge specifications and one nonlearned baseline, not 18 trained models.

No projection or clipping is applied to the reported predictions. This preserves model failures outside the source mechanism's support. It also means that a numerical regression prediction should not be treated as a feasible state estimate merely because its average error is small.

\section{Evaluation Design and Audits}
\subsection{Origins, targets, and score}
One-step evaluation uses all 48 records per route with the contemporaneous observed state. Recursive evaluation uses seven common zero-based origins $\{0,6,12,18,24,30,36\}$ and a 12-step horizon. At each origin, only the observed state prefix through that origin initializes the predictor. Every subsequent state, including states used in updated history summaries, is a prediction. Recorded future plans and deterministic future context remain supplied. All specifications use identical origins and endpoint records.

\begin{figure}[htbp]
\centering
\begin{minipage}{.49\linewidth}\centering\begin{tikzpicture}
\begin{axis}[width=\linewidth,height=49mm,title={All 11 targets},xlabel={Forecast offset (months)},ylabel={nRMSE},xtick={1,3,6,9,12},grid=major,legend style={font=\tiny,at={(.02,.98)},anchor=north west,draw=none},tick label style={font=\scriptsize},label style={font=\scriptsize}]
\addplot+[color=StudyGray,mark=*,solid,mark size=1.5pt,line width=.8pt] coordinates {(1,0.0599362854) (2,0.0903185295) (3,0.113157957) (4,0.130756433) (5,0.146996471) (6,0.159233791) (7,0.168533437) (8,0.17624612) (9,0.182266214) (10,0.185903971) (11,0.188418989) (12,0.191085042)};
\addlegendentry{Current}
\addplot+[color=StudyBlue,mark=*,solid,mark size=1.5pt,line width=.8pt] coordinates {(1,0.0584623616) (2,0.0870969673) (3,0.108491573) (4,0.124692056) (5,0.140156274) (6,0.151521652) (7,0.160434511) (8,0.167939661) (9,0.173685949) (10,0.176809738) (11,0.178885943) (12,0.180771339)};
\addlegendentry{History-6}
\addplot+[color=StudyOrange,mark=*,solid,mark size=1.5pt,line width=.8pt] coordinates {(1,0.0590962513) (2,0.0879970212) (3,0.109251785) (4,0.125239751) (5,0.140639349) (6,0.152416894) (7,0.16159218) (8,0.16946138) (9,0.175187275) (10,0.178338045) (11,0.181237746) (12,0.184461357)};
\addlegendentry{Count-6}
\addplot+[color=StudyTeal,mark=*,solid,mark size=1.5pt,line width=.8pt] coordinates {(1,0.0588770633) (2,0.0882301839) (3,0.110274242) (4,0.126904241) (5,0.142050392) (6,0.153005906) (7,0.161593428) (8,0.168604135) (9,0.173814787) (10,0.17683395) (11,0.178764466) (12,0.180878503)};
\addlegendentry{All history}
\addplot+[color=StudyGray,mark=none,dashed,mark size=1.5pt,line width=.8pt] coordinates {(1,0.0599362854) (2,0.058845269) (3,0.0569617805) (4,0.054715174) (5,0.058673609) (6,0.0562085804) (7,0.0513814701) (8,0.0501920455) (9,0.0487062787) (10,0.046972568) (11,0.0627285071) (12,0.0584891324)};
\addplot+[color=StudyBlue,mark=none,dashed,mark size=1.5pt,line width=.8pt] coordinates {(1,0.0584623616) (2,0.0560816344) (3,0.0545897529) (4,0.0527692286) (5,0.0569659193) (6,0.0547764423) (7,0.0504586957) (8,0.04909553) (9,0.0477669148) (10,0.0462540632) (11,0.0623407993) (12,0.0576450702)};
\end{axis}\end{tikzpicture}\end{minipage}
\begin{minipage}{.49\linewidth}\centering\begin{tikzpicture}
\begin{axis}[width=\linewidth,height=49mm,title={Nine preparation indices},xlabel={Forecast offset (months)},ylabel={nRMSE},xtick={1,3,6,9,12},grid=major,legend style={font=\tiny,at={(.02,.98)},anchor=north west,draw=none},tick label style={font=\scriptsize},label style={font=\scriptsize}]
\addplot+[color=StudyGray,mark=*,solid,mark size=1.5pt,line width=.8pt] coordinates {(1,0.00634279089) (2,0.0108792126) (3,0.0151986077) (4,0.0192079331) (5,0.0249415306) (6,0.0319879542) (7,0.0357283116) (8,0.0395104814) (9,0.043316939) (10,0.0467835192) (11,0.0515937504) (12,0.0579680235)};
\addlegendentry{Current}
\addplot+[color=StudyBlue,mark=*,solid,mark size=1.5pt,line width=.8pt] coordinates {(1,0.0054427048) (2,0.00878320498) (3,0.0118916626) (4,0.0148175943) (5,0.0192366034) (6,0.0237044132) (7,0.0260119109) (8,0.0284448742) (9,0.0309786873) (10,0.033294352) (11,0.0367237313) (12,0.0408549452)};
\addlegendentry{History-6}
\addplot+[color=StudyOrange,mark=*,solid,mark size=1.5pt,line width=.8pt] coordinates {(1,0.00630239942) (2,0.010792869) (3,0.0150744583) (4,0.0190445274) (5,0.0247315543) (6,0.0317448686) (7,0.0354603819) (8,0.0392194243) (9,0.0430052164) (10,0.0464500554) (11,0.0512543781) (12,0.0575775828)};
\addlegendentry{Count-6}
\addplot+[color=StudyTeal,mark=*,solid,mark size=1.5pt,line width=.8pt] coordinates {(1,0.00532445307) (2,0.00851109887) (3,0.0114383207) (4,0.0141965104) (5,0.0182880469) (6,0.0223941071) (7,0.024420091) (8,0.0265830112) (9,0.0288485348) (10,0.0309597584) (11,0.0341121459) (12,0.0376641828)};
\addlegendentry{All history}
\addplot+[color=StudyGray,mark=none,dashed,mark size=1.5pt,line width=.8pt] coordinates {(1,0.00634279089) (2,0.00619564327) (3,0.00609559501) (4,0.00577496651) (5,0.0107314366) (6,0.0107845591) (7,0.00644791643) (8,0.00630731663) (9,0.00623462782) (10,0.00571229698) (11,0.011289178) (12,0.010894062)};
\addplot+[color=StudyBlue,mark=none,dashed,mark size=1.5pt,line width=.8pt] coordinates {(1,0.0054427048) (2,0.00512486888) (3,0.00497556487) (4,0.00469508324) (5,0.0099214786) (6,0.00917022149) (7,0.00542891689) (8,0.00525101181) (9,0.0051432494) (10,0.0047021261) (11,0.0104887478) (12,0.00934355863)};
\end{axis}\end{tikzpicture}\end{minipage}
\caption{Recursive errors with future recorded plans supplied. Dashed lines are matched one-step predictions using contemporaneous observed states at exactly the same endpoint months and origins (current and history-6). The all-record one-step table uses a different record composition.}
\label{fig:curves}
\end{figure}
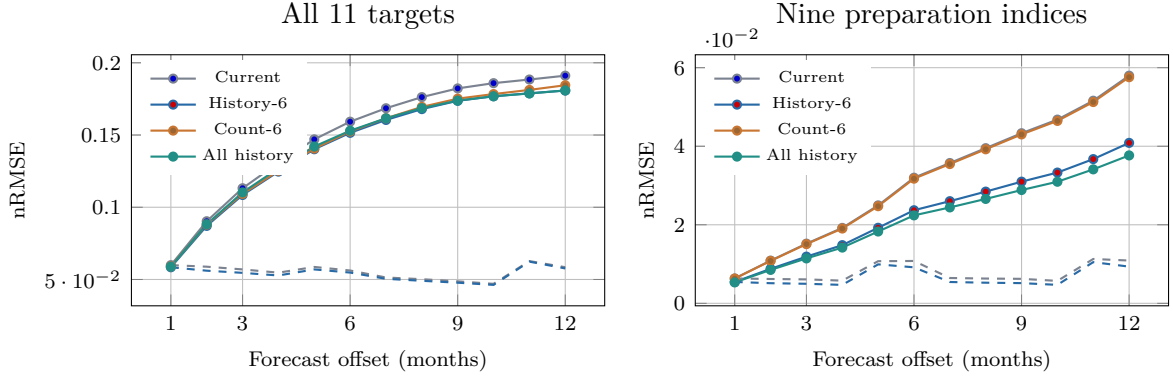

A matched teacher-forced reference predicts the same endpoint record from its contemporaneous true state. This distinguishes recursive error accumulation from changing calendar composition. The all-record one-step score is not substituted for the horizon-one score on the seven-origin subset. Curves in Figure~\ref{fig:curves} should be read under these matched-origin contracts.

For coordinate $d$, let $\sigma_d$ denote its post-state standard deviation in the training set. The aggregate score is
\begin{equation}
 \operatorname{nRMSE}=\sqrt{\frac{1}{11N}\sum_{i=1}^{N}\sum_{d=1}^{11}
 \frac{(\widehat y_{id}-y_{id})^2}{\sigma_d^2}}.
\end{equation}
We also retain coordinate-wise RMSE and MAE. Group scores separate the nine preparation indices, fatigue, and funds. Normalization removes unit differences but does not assign educational utility or ensure equal error contributions.

Paired percentile intervals resample families jointly across compared predictors, preserving every route and profile within a sampled family. We use 10,000 replicates with 40 or 30 clusters, respectively. All families have 125 routes, so equally weighted family MSE reproduces the pooled point score before taking its square root. These intervals are conditional on fixed fitted models and the simulator evaluation distribution. They do not include uncertainty about the realism of the source equations. Secondary windows, components, outcomes, and horizons are reported with descriptive intervals without multiplicity adjustment; we do not select a winning specification from those intervals as a new primary hypothesis.

\subsection{Local freezing and prior exposure}
The expanded protocol specifies the six-month-history versus baseline comparison at one step and the six-step endpoint. A pre-evaluation addendum supplies history-length controls and matched teacher-forced references. These files are locally frozen, hash-recorded analysis specifications; they are not a formally registered preregistration. Earlier pilot results informed the study question.

A subsequent provenance check identified prior exposure of 22 original evaluation families across dynamics, stress, action-fork, and supervised-fine-tuning tasks. The original 40-family results are retained rather than selectively replaced. Exclusion analyses are labeled post-observation sensitivities. A further protocol selected 30 families from the source test partition, excluding all identified prior task families and the entire original 40-family set. It locked original weights, preprocessing, and penalty choices before evaluating those families. This fresh same-rule evaluation includes nine selected learned Ridge specifications and persistence; it is not a new training run, a different simulator, or an external real-student cohort.

\section{Results}
\subsection{History compared with the conventional baseline}
Table~\ref{tab:main} reports the original evaluation. Six-month history lowers one-step nRMSE from 0.058750 to 0.057009, a 2.96\% relative reduction. At the six-step endpoint, the score decreases from 0.159234 to 0.151522, a 4.84\% reduction; the paired difference is $-0.007712$, with interval $[-0.008996,-0.006443]$. Twelve-step scores are 0.191085 and 0.180771. Persistence is substantially weaker, reaching 0.320327 at six steps and 0.526675 at twelve, but this weak baseline alone does not establish the specific value of historical summaries.

Locked evaluation on the fresh families gives a similar baseline comparison (Table~\ref{tab:fresh}). One-step scores are 0.058400 and 0.056828, a 2.69\% relative reduction. Six-step scores are 0.156480 and 0.150074, a 4.09\% reduction, with paired difference $-0.006406$ and interval $[-0.007866,-0.005093]$. These observations extend the pattern to documented previously unused families under identical source rules. They do not test transfer to other mechanisms or actual student records.

\begin{table}[htbp]
\centering
\small
\caption{Original 40-family evaluation: aggregate normalized RMSE. One-step prediction uses all records; recursive endpoints use the same seven origins per route. All predictions remain unprojected.}
\label{tab:main}
\begin{tabular}{lrrr}
\toprule
Specification & 1 step & 6 steps & 12 steps \\
\midrule
Current state + plan & 0.058750 & 0.159234 & 0.191085 \\
History-1 & 0.057834 & 0.156365 & 0.187299 \\
History-3 & 0.057238 & 0.153452 & 0.183228 \\
History-6 & 0.057009 & 0.151522 & 0.180771 \\
History-12 & 0.056945 & 0.150763 & 0.178771 \\
All history & 0.057132 & 0.153006 & 0.180879 \\
Count-6 only & 0.058017 & 0.152417 & 0.184461 \\
Count-all only & 0.058639 & 0.157927 & 0.189845 \\
Current, no plan & 0.064450 & 0.186877 & 0.250815 \\
History-6, no current plan & 0.061598 & 0.174394 & 0.239969 \\
All history, no current plan & 0.061853 & 0.191677 & 0.373956 \\
Persistence & 0.077200 & 0.320327 & 0.526675 \\
\bottomrule
\end{tabular}
\end{table}

\begin{table}[htbp]
\centering
\small
\caption{Fixed-weight validation on 30 previously unused families under identical simulator rules. No preprocessing, coefficient, or penalty choice is fitted to this set.}
\label{tab:fresh}
\begin{tabular}{lrrr}
\toprule
Specification & 1 step & 6 steps & 12 steps \\
\midrule
Current state + plan & 0.058400 & 0.156480 & 0.186687 \\
History-6 & 0.056828 & 0.150074 & 0.176040 \\
History-12 & 0.056719 & 0.149420 & 0.174548 \\
All history & 0.056830 & 0.150692 & 0.175504 \\
Count-6 only & 0.057741 & 0.150632 & 0.180099 \\
Count-all only & 0.058286 & 0.155242 & 0.185422 \\
Current, no plan & 0.063900 & 0.183541 & 0.245420 \\
History-6, no current plan & 0.061259 & 0.172388 & 0.234944 \\
All history, no current plan & 0.061387 & 0.189648 & 0.373426 \\
Persistence & 0.077709 & 0.326292 & 0.543952 \\
\bottomrule
\end{tabular}
\end{table}

\subsection{The history-length control changes the interpretation}
At six steps, the original length-only control scores 0.152417, close to the full six-month-history score of 0.151522. The remaining relative reduction is 0.59\%, and its paired difference $-0.000895$ has interval $[-0.001957,0.000248]$. Fresh-family scores are 0.150632 and 0.150074; the remaining reduction is 0.37\%, with difference $-0.000558$ and interval $[-0.001783,0.000728]$. Table~\ref{tab:effects} separates these comparisons. Both six-step intervals include zero, so the evidence does not resolve an aggregate improvement over the length-only control at this endpoint.

This is not evidence that history has no predictive value. Against the length-only control, one-step differences are $-0.001008$ on the original families and $-0.000913$ on fresh families; their descriptive intervals lie below zero. Twelve-step differences are $-0.003690$ and $-0.004059$, also with intervals below zero. The horizon dependence and target decomposition matter. Nor is the longest summary uniformly best: the original all-history six-step score is 0.153006, compared with 0.150763 for the 12-month summary; fresh scores are 0.150692 and 0.149420. These are fixed-window observations, not a test-selected recommendation of an optimal window.

The all-history model also improves on its own count-all control at six steps: the original difference is $-0.004921$, with descriptive interval $[-0.006202,-0.003696]$, and the fresh-family difference is $-0.004550$, with interval $[-0.005387,-0.003731]$. These controls have different bases. Count-6 introduces a hinge that saturates after six observations, whereas count-all introduces an unsaturated linear count and the initial-record indicator. Consequently, the uncertain six-month comparison must not be generalized to every historical representation or calendar control.

\begin{table}[htbp]
\centering
\small
\caption{Paired family-bootstrap differences against length-only controls. Negative differences favor historical content. These 95\% intervals are descriptive and unadjusted across specifications and horizons.}
\label{tab:effects}
\begin{tabular}{llrrr}
\toprule
Set & Comparison & Step & $\Delta$ nRMSE & 95\% interval \\
\midrule
Original & History-6 vs count-6 & 1 & -0.001008 & [-0.001138, -0.000882] \\
Original & History-6 vs count-6 & 6 & -0.000895 & [-0.001957, +0.000248] \\
Original & History-6 vs count-6 & 12 & -0.003690 & [-0.006299, -0.000782] \\
Original & All history vs count-all & 1 & -0.001507 & [-0.001667, -0.001358] \\
Original & All history vs count-all & 6 & -0.004921 & [-0.006202, -0.003696] \\
Original & All history vs count-all & 12 & -0.008966 & [-0.011098, -0.006899] \\
Fresh & History-6 vs count-6 & 1 & -0.000913 & [-0.001021, -0.000805] \\
Fresh & History-6 vs count-6 & 6 & -0.000558 & [-0.001783, +0.000728] \\
Fresh & History-6 vs count-6 & 12 & -0.004059 & [-0.007459, -0.000661] \\
Fresh & All history vs count-all & 1 & -0.001455 & [-0.001624, -0.001299] \\
Fresh & All history vs count-all & 6 & -0.004550 & [-0.005387, -0.003731] \\
Fresh & All history vs count-all & 12 & -0.009917 & [-0.011729, -0.008024] \\
\bottomrule
\end{tabular}
\end{table}

\subsection{Different state groups give different answers}
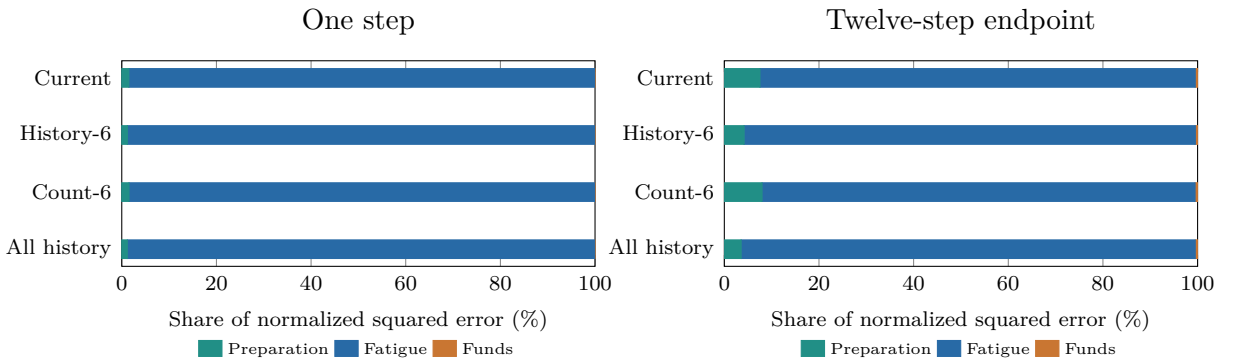
\begin{figure}[htbp]
\centering
\begin{minipage}{.49\linewidth}\centering\begin{tikzpicture}
\begin{axis}[width=\linewidth,height=43mm,title={One step},xbar stacked,xmin=0,xmax=100,ytick={1,2,3,4},yticklabels={Current,History-6,Count-6,All history},y dir=reverse,xlabel={Share of normalized squared error (\%)},bar width=7pt,tick label style={font=\scriptsize},label style={font=\scriptsize},legend style={font=\tiny,at={(.5,-.32)},anchor=north,legend columns=3,draw=none}]
\addplot+[fill=StudyTeal,draw=StudyTeal] coordinates {(1.5018082,1) (1.198402,2) (1.5305506,3) (1.1732533,4)};
\addlegendentry{Preparation}
\addplot+[fill=StudyBlue,draw=StudyBlue] coordinates {(98.408858,1) (98.7166,2) (98.377858,3) (98.743859,4)};
\addlegendentry{Fatigue}
\addplot+[fill=StudyOrange,draw=StudyOrange] coordinates {(0.089333618,1) (0.084998451,2) (0.091591358,3) (0.082887987,4)};
\addlegendentry{Funds}
\end{axis}\end{tikzpicture}\end{minipage}
\begin{minipage}{.49\linewidth}\centering\begin{tikzpicture}
\begin{axis}[width=\linewidth,height=43mm,title={Twelve-step endpoint},xbar stacked,xmin=0,xmax=100,ytick={1,2,3,4},yticklabels={Current,History-6,Count-6,All history},y dir=reverse,xlabel={Share of normalized squared error (\%)},bar width=7pt,tick label style={font=\scriptsize},label style={font=\scriptsize},legend style={font=\tiny,at={(.5,-.32)},anchor=north,legend columns=3,draw=none}]
\addplot+[fill=StudyTeal,draw=StudyTeal] coordinates {(7.5296263,1) (4.1790728,2) (7.9716071,3) (3.5475867,4)};
\addlegendentry{Preparation}
\addplot+[fill=StudyBlue,draw=StudyBlue] coordinates {(91.972686,1) (95.341236,2) (91.494432,3) (95.953545,4)};
\addlegendentry{Fatigue}
\addplot+[fill=StudyOrange,draw=StudyOrange] coordinates {(0.49768793,1) (0.47969094,2) (0.53396054,3) (0.49886861,4)};
\addlegendentry{Funds}
\end{axis}\end{tikzpicture}\end{minipage}
\caption{Additive shares of the aggregate training-standardized squared error. Fatigue dominates the aggregate despite normalization; improvements in other coordinates can therefore be obscured. RMSE itself is not additive.}
\label{fig:composition}
\end{figure}

Figure~\ref{fig:composition} decomposes the original evaluation's normalized squared error. For the six-month-history predictor, fatigue contributes 98.72\% of one-step loss and 97.80\% of six-step loss. The nine preparation coordinates contribute only 1.20\% and 2.00\%, respectively. This reflects their differing predictability relative to training variation, rather than an intentional preference for fatigue.

An independently replayed exploratory comparison against the length-only control finds a six-step preparation-group nRMSE difference of $-0.008040$, with paired interval $[-0.008538,-0.007556]$; all 40 families favor the historical summaries on that group score. Funds give $-0.002280$, with interval $[-0.002631,-0.001939]$ and 39 family wins. Fatigue instead gives $+0.001135$, with interval $[-0.002419,0.005009]$. The aggregate comparison therefore combines clearer preparation and funds improvements with uncertainty in its dominant coordinate. A claim of uniform improvement across the student state would misrepresent this result. These group intervals are exploratory and derived from the original 40 families; they are not presented as independent real-world validation.

\begin{table}[htbp]
\centering
\small
\caption{Exploratory six-step group comparison: history-6 minus count-6 on the original 40 families. Group scores use their own coordinate averages; they are not additive RMSE contributions.}
\label{tab:groups}
\begin{tabular}{lrrr}
\toprule
State group & $\Delta$ group nRMSE & 95\% interval & Family wins \\
\midrule
Preparation (9) & -0.008040 & [-0.008538, -0.007556] & 40/40 \\
Fatigue & +0.001135 & [-0.002419, +0.005009] & 22/40 \\
Funds & -0.002280 & [-0.002631, -0.001939] & 39/40 \\
\bottomrule
\end{tabular}
\end{table}

\subsection{Predictive error does not ensure valid support}
The source fatigue transition clamps outcomes to $[5,90]$. A broad $[0,100]$ check misses many infeasible predictions. Table~\ref{tab:bounds} retains both checks. On the original evaluation, the baseline violates the strict interval in 31.369\% of one-step and 30.786\% of six-step predictions. The six-month-history model reduces these rates to 28.860\% and 28.200\%, but its six-step broad-interval violation rate is only 0.123\%. All of its strict six-step violations are predictions below 5.

These results preclude describing the unprojected outputs as consistently support-valid. Projection onto a verified target interval cannot increase pointwise squared error when the target belongs to that interval; such an improvement would be an elementary constraint effect rather than a new learning contribution. Recursive projection would change later inputs and require a separate evaluation. We preserve the original unprojected predictions here rather than silently repairing the principal results.

\begin{table}[htbp]
\centering
\small
\caption{Original evaluation fatigue predictions outside the actual [5,90] support versus the broader [0,100] interval. Strict failures are retained in the reported scores.}
\label{tab:bounds}
\begin{tabular}{lrrrr}
\toprule
Specification & Strict: 1 & Strict: 6 & Strict: 12 & Wide: 6 \\
\midrule
Current state + plan & 31.369\% & 30.786\% & 31.500\% & 0.137\% \\
History-6 & 28.860\% & 28.200\% & 28.560\% & 0.123\% \\
Count-6 only & 30.404\% & 30.414\% & 32.300\% & 0.089\% \\
All history & 28.809\% & 27.280\% & 28.600\% & 0.437\% \\
\bottomrule
\end{tabular}
\end{table}

\section{Reproducibility, Limitations, and Implications}
The reproducibility artifacts record source, data, protocol, result, and checkpoint hashes; exact group manifests; the full penalty grid; coordinate and family errors; and independent replay code. Checks verify split disjointness, state continuity, train-only preprocessing, and locked fresh-family weights. Independent inference without importing the training implementation reproduces all five reported task scores for all 17 learned specifications exactly at the recorded numerical precision. Matched-origin teacher-forced versus horizon-one recursive predictions differ by at most $8.54\times10^{-8}$ in training-standardized units. Poisoning unobserved future states and target labels leaves permitted-origin predictions unchanged. These checks establish implementation consistency for the declared conditional contract, not empirical educational validity.

Several limitations define the scope. First, the mechanism and numeric coefficients are designed assumptions; multiple simulated routes cannot replace authorized longitudinal observations. Second, current and future saved plans can encode event information. Third, only transparent linear summary representations are tested, so conclusions neither characterize all recurrent systems nor evaluate language-model conversational memory. Fourth, prior exposure affected the original family set, although locked fresh-family evaluation addresses the documented overlap within the same mechanism. Fifth, conditional family-bootstrap intervals and unadjusted secondary intervals should not be interpreted as universal or simultaneous guarantees. Finally, prediction gains do not measure policy benefits: no autonomous action-selection claim follows from this experiment.

The methodological implication is practical. Evaluations of history should report the prediction-time information contract, add clock-aware controls when history handling creates calendar features, align recursive and teacher-forced targets, preserve dependence among scenario groups, and inspect each consequential output rather than rely on one aggregate. Here those steps yield a qualified result: compact historical summaries contain useful predictive information for several simulator coordinates, while calendar controls and fatigue error substantially change the interpretation of the aggregate six-step gain. A credible extension to university planning would require independently measured outcomes and a policy evaluation against strong fixed and expert baselines, with the same attention to information timing and failure disclosure.

\clearpage
\appendix
\section{Complete Specifications and Artifact Contract}
The accompanying computational material contains deterministic reconstruction code, frozen protocols and amendments, split manifests, fitted model arrays, validation penalty candidates, per-family errors, and independent replay audits. The source generator and its required factual-reference pool must remain frozen to reproduce the archived scenario profiles. No archival data license or individual source permission is inferred from public availability. Until a permanent public repository is supplied, this manuscript should be treated as a local reproducibility package rather than a claim of public open-source availability.

The declared information boundary excludes identifiers, route-policy labels, hidden generator assumptions, actual hours, realized spending, completion status, event fields, target outcomes, and post-execution narrative fields. In a recursive forecast, historical states after the origin are model predictions. Saved future plans are nevertheless permitted conditioning variables and may encode subsequent realized information.

\section{AI Tools and Human Responsibility}
Generative AI tools assisted research organization, code development, literature screening, drafting, and independent code and result checks. Statistical tables and figures were computed from recorded experiment outputs; they were not generated as plausible illustrations of an unrun experiment. A human submitting author must verify every claim, citation, provenance statement, and permission and assumes responsibility for the final work. AI tools are not authors. No human author, affiliation, ethical approval, observational cohort, or data-sharing permission has been invented for this draft.

\begin{table}[htbp]
\centering
\small
\caption{Complete original specifications. Component-only and component-removal rows retain current state, context, current plan, and history-count indicators; only the named past-summary component changes. Explicit current-plan removal rows are separate. All results, including unfavorable outcomes, are retained. Penalties were selected on validation data only.}
\label{tab:all}
\begin{tabular}{lrrr}
\toprule
Specification & 1 step & 6 steps & 12 steps \\
\midrule
Current state + plan & 0.058750 & 0.159234 & 0.191085 \\
History-1 & 0.057834 & 0.156365 & 0.187299 \\
History-3 & 0.057238 & 0.153452 & 0.183228 \\
History-6 & 0.057009 & 0.151522 & 0.180771 \\
History-12 & 0.056945 & 0.150763 & 0.178771 \\
All history & 0.057132 & 0.153006 & 0.180879 \\
History-6: past-state mean only & 0.057615 & 0.152877 & 0.185093 \\
History-6: past slope only & 0.057328 & 0.151982 & 0.183769 \\
History-6: past-plan mean only & 0.057945 & 0.152914 & 0.184963 \\
History-6: omit past-state mean & 0.057008 & 0.151687 & 0.181453 \\
History-6: omit past slope & 0.057503 & 0.153343 & 0.185511 \\
History-6: omit past-plan mean & 0.057330 & 0.151891 & 0.183382 \\
History-6, no current plan & 0.061598 & 0.174394 & 0.239969 \\
All history, no current plan & 0.061853 & 0.191677 & 0.373956 \\
Current, no plan & 0.064450 & 0.186877 & 0.250815 \\
Count-6 only & 0.058017 & 0.152417 & 0.184461 \\
Count-all only & 0.058639 & 0.157927 & 0.189845 \\
Persistence & 0.077200 & 0.320327 & 0.526675 \\
\bottomrule
\end{tabular}
\end{table}

\end{document}